%% file: main.tex
\documentclass[a4paper,fleqn]{cas-dc}

\usepackage[numbers,square,sort&compress]{natbib}
\usepackage{amsmath,amssymb,amsfonts,amsthm}
\usepackage{graphicx}
\usepackage{booktabs}
\usepackage{tabularx}
\usepackage{array}
\usepackage{multirow}
\usepackage{makecell}
\usepackage{microtype}
\usepackage{xcolor}
\usepackage{url}
\usepackage{caption}
\usepackage{subcaption}
\usepackage{enumitem}

\def\tsc#1{\csdef{#1}{\textsc{\lowercase{#1}}\xspace}}
\tsc{WGM}
\tsc{QE}
\tsc{EP}
\tsc{PMS}
\tsc{BEC}
\tsc{DE}

\begin{document}
\let\WriteBookmarks\relax
\def\floatpagepagefraction{1}
\def\textpagefraction{.001}

\shorttitle{LENS-GRF: Structured Lesion-Set Evidence for Gated Global-Prior Refinement}

\shortauthors{M. M. Shahriar et~al.}

\title [mode = title]{LENS-GRF: Permutation-Invariant Lesion Evidence Network with Gated Residual Fusion for Acne Severity Grading and Multi-Rater Clinical Oracle Analysis}                      

\author[1]{Muhammad Muhtasim Shahriar}[orcid=0009-0008-4954-3811]


\ead{c221186@ugrad.iiuc.ac.bd}

\credit{Conceptualization, Methodology, Software, Validation, Formal analysis, Investigation, Data Curation, Writing - Original Draft, Writing - Review \& Editing, Visualization, Project administration}

\affiliation[1]{organization={Department of Computer Science, International Islamic University Chittagong (IIUC)},
    city={Chittagong},
    country={Bangladesh}}

\author[2]{M. F. Mridha}[orcid=0000-0001-5738-1631]
\cormark[1]
\ead{firoz.mridha@aiub.edu}
\affiliation[2]{organization={Department of Computer Science, American International University-Bangladesh (AIUB)},
    city={Dhaka},
    country={Bangladesh}}

\author[3]{Md. Rakibul Islam}
\ead{md.rakibul.islam@mdu.se}
\affiliation[3]{
    organization={Department of Computer Science and Engineering, Mälardalen University},
    city={Västerås},
    state={Västmanland},
    country={Sweden}
}

\cortext[cor1]{Corresponding author}


\begin{abstract}
Localized lesion evidence may improve automated acne severity grading, but detector errors and differences in clinical grading criteria can limit generalization. We evaluated LENS-GRF, a lesion evidence network with permutation-invariant aggregation and gated residual fusion that combines a global Vision Transformer with lesion-region embeddings derived from YOLO11s proposals. Fully automated four-class grading was evaluated on 292 held-out ACNE04 images and, without external adaptation, on 200 PLSBRACNE01 participants represented by 600 facial views. Annotation-assisted inference used registered lesion boxes from three dermatologists in a matched subset of 148 participants. Performance measures included accuracy, macro-averaged precision, sensitivity, and one-versus-rest specificity. Paired comparisons used exact McNemar tests with Holm adjustment. On ACNE04, automated LENS-GRF achieved 80.82\% accuracy, 78.80\% macro precision, 78.92\% sensitivity, and 92.78\% specificity. With supplied lesion boxes, internal accuracy averaged 95.89\% across three seeds. On the full external cohort, automated LENS-GRF achieved 35.00\% accuracy, compared with 42.50\% for an independently trained global baseline. External macro precision, sensitivity, and specificity for LENS-GRF were 50.52\%, 34.31\%, and 78.19\%, respectively. In the matched subset, accuracy using Doctor A's lesion annotations was 47.97\%, compared with 35.14\% for automated inference (unadjusted $p=0.0094$; Holm-adjusted $p=0.0846$). Strong internal performance with supplied lesion annotations did not translate into reliable automated grading on the external cohort. More reliable lesion detection, modelling that accounts for differences in grading criteria, and independent validation are needed before clinical use.
\end{abstract}


\maketitle



\section{Introduction}

Acne vulgaris is a common inflammatory skin condition. Reproducible severity assessment supports clinical monitoring, treatment decisions, and outcome measurement. A systematic review of artificial intelligence for acne assessment identifies subjectivity and the time required for conventional global grading and manual lesion counting as practical limitations~\citep{lensintroTraini2025}. Clinical grading systems also differ in the weight they assign to lesion counts, inflammatory morphology, and anatomical distribution. A study of acne grading across clinical standards identified inconsistent diagnostic criteria, ambiguity in lesion assessment, and limited annotations as continuing challenges~\citep{lensintroChen2025}.

These differences complicate the evaluation of a model trained under one grading criterion and tested under another. In this study, the internal ACNE04 benchmark uses count-oriented classes related to the Hayashi criterion, while the external PLSBRACNE01 benchmark uses assessments related to the Pillsbury system. Although both benchmarks contain four ordinal categories, those categories should not be treated as clinically interchangeable. Evaluation across the two datasets therefore involves changes in both image characteristics and grading criteria. Both must be considered when interpreting negative transfer results and apparent improvements obtained by supplying more reliable lesion annotations.

Whole-face appearance and local lesion characteristics provide complementary information for acne grading. Image-level representations capture regional skin context, but resizing images to standard input dimensions may make small lesions less visible. Diffuse redness, lesion clusters, and isolated inflammatory structures also occur at different spatial scales. A comparison of SqueezeNet, CenterNet, and VGG-16 for Hayashi-based severity classification showed that model choice affected grading performance within the evaluated dataset and protocol~\citep{lensintroShaik2025}. The AcneGrader approach uses ensemble pruning to remove redundant deep classifiers while retaining image-level grading capability~\citep{lensintroLiu2022}. A controlled study of SimCLR-based self-supervised learning under limited annotation availability found that supervised transfer learning generally performed better in the acne-grading settings tested~\citep{lensintroSrijiranon2026}.

These studies examine image representations, but their experiments do not establish whether a variable number of local lesion candidates can reliably refine a global prediction. A method that combines whole-face and lesion-level information must therefore account for how errors in local detections affect an otherwise useful global estimate.

Other methods address uncertainty between severity grades or use localized lesion evidence. Label-distribution smoothing has been investigated on ACNE04 to represent neighboring grades more continuously than conventional hard labels~\citep{lensintroProkhorov2024}. For smartphone photographs, a Faster R-CNN lesion detector was combined with a separate LightGBM severity grader, treating localization and image-level grading as separate tasks~\citep{lensintroHuynh2022}. An evaluation of object detectors for acne localization and count-based severity estimation identified small lesion size as a limitation of general-purpose detectors~\citep{lensintroWen2022}. A comparison of region-based detectors and YOLO models for acne identification provided detection results, but these do not establish agreement in patient-level severity grading~\citep{lensintroTrivedi2024}.

These studies use global or local evidence under different datasets, grading definitions, and evaluation units. Lesion-detection precision and image- or subject-level severity accuracy measure different outcomes, so results from unmatched benchmarks cannot support a direct ranking of methods. Review literature also identifies limited independent and prospective evaluation as a barrier to clinical translation~\citep{lensintroTraini2025}. These gaps motivate evaluating whether an unordered set of lesion observations can refine a global prediction and under which changes in imaging conditions or grading criteria that refinement becomes unreliable.

We investigate this question through LENS-GRF, a multistage framework for four-class photographic acne severity grading. A global Vision Transformer branch produces a whole-image representation and baseline severity logits. The automatic lesion branch uses YOLO11s to detect candidate regions, encodes the appearance and geometry of the retained regions, and aggregates their tokens through an attention-based set representation. This aggregation is invariant to the ordering of the selected valid tokens. It does not make detector thresholding or top-$K$ selection invariant to changes in how upstream candidates are constructed.

A scalar gate controls the residual contribution of local evidence to the global prediction. When the gate is zero, the fused logits equal the global logits exactly. This conditional identity alone does not establish that the learned gate will suppress inaccurate detections under domain shift.

We evaluate inference using predicted lesion boxes separately from annotation-assisted oracle inference using supplied boxes. The experiments include internal evaluation on ACNE04, a frozen cross-dataset assessment on PLSBRACNE01, and a matched sensitivity analysis using lesion annotations from three dermatologists. The sensitivity analysis measures variation in model outputs across annotation sources; it does not measure direct agreement among the dermatologists' image-level severity diagnoses. Together, these evaluations assess the usefulness of structured lesion evidence and its sensitivity to upstream detection errors. Strong internal oracle performance is therefore examined alongside the limits of automatic cross-dataset transfer.

This paper makes the following contributions:
\begin{enumerate}
    \item A lesion-evidence representation that encodes the appearance and geometry of a variable number of detected regions and uses permutation-invariant attention to aggregate the selected valid lesion tokens.

    \item A gated residual-logit formulation that combines a global image prediction with local lesion evidence. The formulation has an explicit identity condition when the gate is zero, although this property does not establish automatic protection against errors under distribution shift.

    \item An internal ACNE04 evaluation that distinguishes fully automated detection from annotation-assisted oracle inputs, together with a split-integrity audit that informs the interpretation of benchmark results.

    \item A frozen PLSBRACNE01 evaluation that documents the limitations of automatic cross-dataset transfer and examines detector scale, candidate saturation, and gate behavior as possible contributors to the observed failure.

    \item A matched oracle sensitivity analysis using registered lesion boxes from three dermatologists to measure variation in model performance and predictions across annotation sources.
\end{enumerate}

The remainder of this paper is organized as follows. Section~1.1 reviews closely related work on acne grading, lesion detection, and evidence fusion. Section~2 describes the methods and validation protocols. Section~3 reports the results, Section~4 discusses the findings and limitations, and Section~5 presents the conclusions.




\subsection{Related Work}

\subsubsection{Whole-Face Acne Severity Grading and Ordinal Modeling}

Computational acne grading draws on both overall facial appearance and lesion burden. The ACNE04 benchmark was introduced alongside a joint image-grading and lesion-counting framework based on label-distribution learning~\citep{lensrwWu2019}. The framework represented grading and count targets as distributions to account for uncertainty between neighboring grades and lesion burdens. This approach supports count-sensitive ordinal prediction, but its evaluation was confined to the ACNE04 grading protocol and did not establish equivalence with other clinical criteria. It also addressed a different task from determining whether automatically detected lesion regions should modify an existing whole-image prediction.

Other methods combine whole-image classification with lesion localization to provide information at both scales. An ensemble-based acne analysis framework combined separate classification and localization modules~\citep{lensrwZhangMa2022}. The outputs of these modules require interpretation according to their respective evaluation units; localization performance and severity-grading performance do not constitute a single measure of clinical agreement. Another approach extracted lesion-related prior knowledge to guide a grading network~\citep{lensrwLinPrior2022}. The segmentation-derived guidance explicitly incorporated local information, although both the prior representation and the predicted grade remained tied to the training and evaluation protocol.

These studies establish image classification, count supervision, and guided local evidence as approaches to acne grading. A remaining question is how reliably a variable number of lesion proposals can be used at inference and under what conditions that local evidence should modify a global prediction.

\subsubsection{Lesion Detection, Counting, and Local Morphological Cues}

Lesion localization operates at a finer prediction scale than whole-face severity grading. The ACNet framework combined mask-aware attention with dynamic context enhancement to improve acne detection under difficult visual conditions~\citep{lensrwMin2021}. An improved YOLOv7 architecture has also been proposed for automatic lesion detection, focusing on feature representation and localization~\citep{lensrwZhangYOLOv7_2024}. These studies provide evidence about detecting small or heterogeneous lesions, but their detection results do not establish performance in four-level ordinal severity grading.

Mobile applications and limited computing resources introduce additional design constraints. A YOLO-based mobile application has been developed to recognize lesion types, including cysts, papules, pustules, and whiteheads~\citep{lensrwIsa2021}. Lesion-type recognition provides information distinct from an overall severity category. An optimizer comparison for YOLOv8 acne detection reported differences in lesion-level precision, recall, and average precision across training configurations~\citep{lensrwIlahi2025}.

A comparison of YOLOv9 through YOLOv11 on ACNE04 examined preprocessing and augmentation while reporting persistent confusion between adjacent count-derived severity classes~\citep{lensrwKhairani2025}. A modified YOLOv11n detector emphasized computational efficiency and showed weak performance on an ACNE04-derived external test collection, an evaluation setting that differs from subject-level grading across clinical criteria~\citep{lensrwArdiansyah2025}. Skin-color analysis and augmentation have also been incorporated into YOLOv11 for inflammatory-lesion detection~\citep{lensrwNainggolan2025}. That study evaluated localization and inflammatory lesion categories and did not assess frozen transfer of a four-grade severity classifier.

Candidate generation, lesion-type recognition, and severity estimation therefore require separate evaluation. Improving a bounding-box detector may yield better local inputs, but it does not establish that a downstream grading model will remain calibrated when lesion size, image framing, false-positive detections, or the grading rubric change. Limiting the input to a fixed number of boxes can reduce computation, yet relevant evidence may be lost when lesions are numerous or proposals are unreliable. These effects need to be measured in the complete grading pipeline; a detector's average precision alone cannot establish their impact on severity predictions.

\subsubsection{Global--Local Evidence Fusion, Priors, and Structured Lesion Representation}

Prior-guided grading provides a close precedent for explicit lesion-set fusion. Segmentation-derived lesion priors have been used to guide a severity classifier~\citep{lensrwLinPrior2022}. A separate acne-analysis system combined distinct classification and localization components~\citep{lensrwZhangMa2022}. These methods incorporate local evidence, but neither representation is equivalent to an unordered learned set of tokens encoding lesion appearance and geometry. The way local information enters the grading model therefore requires separate methodological comparison.

The KIEGLFN framework combined key-information enhancement with global--local feature fusion~\citep{lensrwLinKIEGLFN2022}. The framework addresses differences between count-based and morphology-oriented clinical criteria and investigates transfer fine-tuning when data for the second criterion are limited. KIEGLFN therefore provides an established example of guided fusion beyond simple feature concatenation. Its adaptation procedure must also be considered when interpreting performance under a different grading criterion: the reported results do not constitute a locked zero-shot evaluation.

Diagnostic Evidence Distillation (DED) uses local information differently~\citep{lensrwLinDED2023}. A teacher model learns from lesion-related diagnostic evidence, including spatial information and count supervision, and transfers this knowledge to an image-only student. The student can then predict severity without requiring the same lesion annotations at inference. This design separates the evidence used during training from the inputs required at deployment. DED reports results on ACNE04 and PLSBRACNE01, but these were obtained through dataset-specific development protocols. They should not be interpreted as an untouched, zero-shot transfer from ACNE04 to PLSBRACNE01.

AcneAI takes a lesion-first approach by segmenting individual acne regions, estimating a score for each lesion, and combining those scores into a continuous severity measure~\citep{lensrwGazeauAcneAI}. Its evaluation examines correlation with ACNE04 severity levels and includes a separate clinical dataset for inflammatory-lesion assessment. This aggregation differs from a learned permutation-invariant attention encoder coupled to global severity logits. The outcome scales and clinical evaluation units also differ and do not support direct comparison.

Previous studies have thus used local information in several forms. The relevant distinctions concern how lesion evidence is represented, whether it is required at inference, and how its contribution to the global prediction is controlled. For a gated residual formulation,
\[
    \mathbf{z}_{f} = \mathbf{z}_{g} + g\,\Delta\mathbf{z},
\]
the fused logits equal the global logits when $g=0$. This conditional identity does not establish that the learned gate will reach zero when detector outputs are unreliable.

\subsubsection{External Evaluation, Clinical-Criteria Shift, and Annotation Variability}

External evaluation requires attention to both image acquisition and the definition of the reference labels. The count-oriented targets used in ACNE04's original joint grading-and-counting study~\citep{lensrwWu2019} differ from the Pillsbury-oriented categories in PLSBRACNE01 discussed in criterion-adaptation research~\citep{lensrwLinKIEGLFN2022}. KIEGLFN investigated transfer fine-tuning between these criteria, while DED evaluated both benchmarks under its own training and testing procedures~\citep{lensrwLinDED2023}. Their reported results cannot be ranked directly against those of a model evaluated on PLSBRACNE01 with weights and thresholds frozen after development on ACNE04. Comparisons must also account for the evaluation unit: an image, a facial view, or a subject represented by aggregated predictions from multiple views.

Other validation settings address different questions. AcneAI evaluated an inflammatory-lesion endpoint using a multicentre clinical image collection and separately examined the relationship between its continuous severity score and ACNE04 labels~\citep{lensrwGazeauAcneAI}. The modified YOLOv11n study evaluated lesion detection on an additional ACNE04-derived collection~\citep{lensrwArdiansyah2025}. Its use of the term \textit{external} therefore does not imply independent evaluation under a different clinical grading system. Changes in image distribution and reference-label definitions can occur together. Without controlled evidence, a decline in performance cannot be attributed exclusively to either source.

Expert annotations introduce another source of variation. Clinicians may supply lesion boxes, overall severity labels, or both, and these forms of reference information are not interchangeable. In LENS-GRF, predictions generated from the separate lesion-box annotations of three dermatologists are compared to measure annotation-conditioned model-output sensitivity. This analysis does not directly measure pairwise agreement among the dermatologists' clinical severity decisions. The evaluation instead asks how a complete, frozen grading pipeline responds to changes in lesion proposals, annotation practices, and clinical criteria. A single annotated dataset cannot be assumed to provide a universal diagnostic reference.

\subsubsection{Synthesis, Methodological Positioning, and Remaining Gap}

The reviewed studies establish precedents for count-aware ordinal learning, detection-based analysis, prior-guided global--local modeling, and diagnostic evidence distillation. Their requirements and outputs differ. Some use lesion supervision during training, others require boxes at inference, and others report continuous severity scores or lesion-localization metrics. Differences in reported accuracy therefore cannot be attributed solely to the fusion module without matching the datasets, reference labels, and evaluation units. Table~\ref{tab:related_work_positioning} summarizes representative approaches while retaining these distinctions and avoiding a ranking based on incomparable metrics.

\begin{table*}[t]
\centering
\caption{Methodological positioning of representative acne-grading and lesion-evidence frameworks.}
\label{tab:related_work_positioning}
\footnotesize
\setlength{\tabcolsep}{3.4pt}
\begin{tabularx}{\textwidth}{@{}>{\raggedright\arraybackslash}p{2.7cm} >{\raggedright\arraybackslash}p{2.0cm} >{\raggedright\arraybackslash}p{2.15cm} >{\raggedright\arraybackslash}p{2.25cm} >{\raggedright\arraybackslash}p{2.6cm} >{\raggedright\arraybackslash}X@{}}
\toprule
Study & Task / label & Global context & Local evidence & Fusion / representation & External evaluation / boundary \\
\midrule
\citet{lensrwWu2019} & ACNE04 grades and counts & CNN grading & Lesion-count targets & Label-distribution learning & Internal ACNE04 evaluation \\
\citet{lensrwZhangMa2022} & Severity and lesion boxes & Classification module & YOLOv5 localization & Linked classifier / localizer & No frozen cross-criterion grading test \\
Lin et~al. (prior-guided)~\citep{lensrwLinPrior2022} & ACNE04 and PLSBRACNE01 grades & Guided classifier & Segmented lesion prior & Prior extraction and guidance & Protocol-specific benchmark evaluation \\
Lin et~al. (KIEGLFN)~\citep{lensrwLinKIEGLFN2022} & Two grading criteria & Global--local network & Key regions and patches & KIE + global--local fusion & Transfer fine-tuning to PLSBRACNE01 \\
Lin et~al. (DED)~\citep{lensrwLinDED2023} & Two grading criteria & Image-only student & Teacher lesion masks / counts & Evidence distillation & Dataset-specific training and tests \\
Gazeau et~al. (AcneAI)~\citep{lensrwGazeauAcneAI} & Continuous severity score & Image-level aggregate & Lesion segmentation / scores & Nonlinear score aggregation & Separate ACNE04 and clinical lesion endpoints \\
LENS-GRF (this study) & Four severity grades & Global ViT logits & Selected ROI tokens & Set attention + gated residual & Frozen external test; automatic transfer limited \\
\bottomrule
\end{tabularx}
\end{table*}

The methodological motivation for LENS-GRF is thus narrower than a claim to introduce global--local acne grading for the first time. It investigates order-invariant aggregation \textit{after} candidate selection, a conditional residual update to a global classifier, and failure-sensitive evaluation when lesion proposals and reference criteria shift. Its external results must be read as evidence about the limitations of that particular automated pipeline, not proof of operational robustness or a clinical safety mechanism. Section 2 specifies the implemented architecture, datasets, evaluation regimes and statistical procedures used to test these questions.




\input{sections/02_methods.tex}

\section{Results}
\label{sec:results}

\subsection{Internal Evaluation, Exploratory Comparisons, and Data Integrity}
\label{sec:internal_experiments}

\subsubsection{ACNE04 Evaluation Cohort}
\label{sec:dataset_protocol}

The archived ACNE04 experimental fold contained 1,457 images: 1,049 for training, 116 for validation, and 292 for testing. The full fold comprised 497 Mild, 637 Moderate, 186 Severe, and 137 Very Severe images, with 99, 131, 33, and 29 images, respectively, in the test partition. The source audit recorded 18,983 lesion annotations. These figures describe the folder labels and annotations used in the experiments; the dataset was not clinically re-annotated. The severity thresholds derive from the half-face inflammatory-lesion criteria of Hayashi and colleagues~\citep{lensresultsHayashi2008}. Raw counts across all lesion-box types do not reproduce that clinical measurement. Section~\ref{sec:methodology} describes dataset selection, preprocessing, the implemented training schedule with a maximum of 55 epochs, gradient accumulation, and metric definitions.

\subsubsection{ACNE04 Benchmark Results}
\label{sec:benchmark_results}

Table~\ref{tab:acne04_benchmarks} reports image-level performance on the 292-image test fold. Except where noted, metrics were independently recomputed from the archived four-class confusion matrices using the fixed class order Mild, Moderate, Severe, and Very Severe. The separately trained ConvNeXtV2-L classifier achieved 77.74\% accuracy. Automatic LENS-GRF, using YOLO11s proposals at a confidence threshold of 0.075 and retaining up to 16 regions, achieved 80.82\% accuracy, a macro-F1 of 0.7877, quadratic weighted kappa (QWK) of 0.8807, balanced accuracy of 78.92\%, and ordinal mean absolute error (MAE) of 0.1952. Its accuracy was 3.08 percentage points higher than that of the independently trained ConvNeXtV2-L classifier. This descriptive comparison does not isolate the contribution of the local branch because the underlying classifiers differ and were trained separately.

\begin{table*}[t]
\centering\small
\caption{ACNE04 image-level results ($N=292$). Metrics were independently reconstructed from archived test confusion matrices. The $\dagger$ row retains values reported in the supplied manuscript because its corresponding prediction matrix has not yet been recovered. A dash indicates an unverified value, not zero. GT denotes supplied ground-truth lesion information. The XGBoost comparison ($\ddagger$) uses GT counts during evaluation and is therefore not an autonomous baseline.}
\label{tab:acne04_benchmarks}
\renewcommand{\arraystretch}{1.12}
\resizebox{\textwidth}{!}{%
\begin{tabular}{@{}llrrrrr@{}}
\toprule
Model/configuration & Lesion input & Acc. (\%) & Macro-F1 & QWK & B-Acc. (\%) & MAE \\
\midrule
\multicolumn{7}{@{}l}{\textit{Independent whole-image models}} \\ \addlinespace[2pt]
Global ConvNeXtV2-L & None & 77.74 & 0.7610 & 0.8506 & 78.47 & 0.2329 \\ 
Global ViT-B/16 (reported)$^{\dagger}$ & None & 78.42 & 0.7695 & 0.8584 & --- & 0.2329 \\ 
\midrule
\multicolumn{7}{@{}l}{\textit{Automatic lesion proposal configurations}} \\ \addlinespace[2pt]
YOLO11s, $\tau=0.150$, $K=16$ & Predicted & 76.71 & 0.7196 & 0.8325 & 69.81 & 0.2397 \\ 
YOLO12s, $\tau=0.075$, $K=16$ & Predicted & 78.42 & 0.7588 & 0.8664 & 77.18 & 0.2226 \\ 
YOLO11s, $\tau=0.075$, $K=32$ & Predicted & 79.79 & 0.7819 & 0.8771 & 79.30 & 0.2055 \\ 
\textbf{LENS-GRF: YOLO11s, $\tau=0.075$, $K=16$} & \textbf{Predicted} & 80.82 & 0.7877 & 0.8807 & 78.92 & 0.1952 \\ 
\midrule
\multicolumn{7}{@{}l}{\textit{Annotation/count-assisted configurations (not autonomous)}} \\ \addlinespace[2pt]
Earlier GT-box oracle & GT boxes & 93.84 & 0.9131 & 0.9627 & 89.83 & 0.0616 \\ 
Oracle ViT-only, seed 42 & GT boxes & 96.23 & 0.9552 & 0.9773 & 95.07 & 0.0377 \\ 
\textbf{Oracle ViT-only, three-seed mean $\pm$ SD} & GT boxes & \textbf{95.89 $\pm$ 0.59} & \textbf{0.9502 $\pm$ 0.0069} & \textbf{0.9753 $\pm$ 0.0033} & \textbf{94.90 $\pm$ 0.24} & \textbf{0.0411 $\pm$ 0.0059} \\ 
XGBoost GT-count oracle$^{\ddagger}$ & GT count & 95.89 & 0.9408 & 0.9758 & 94.56 & 0.0411 \\ 
\bottomrule
\end{tabular}}
\end{table*}

On the same ACNE04 test fold, an annotation-assisted ViT-only LENS-GRF configuration achieved 96.23\% accuracy with seed 42. Across seeds 42, 52, and 62, mean accuracy was $95.89\% \pm 0.59\%$, balanced accuracy was $94.90\% \pm 0.24\%$, macro-F1 was $0.9502 \pm 0.0069$, and QWK was $0.9753 \pm 0.0033$. These results describe performance with supplied lesion boxes. They do not represent detector-driven grading, a theoretical accuracy ceiling, or evidence of autonomous clinical performance. The earlier XGBoost comparison achieved 95.89\% accuracy using ground-truth counts. However, its classifier-fitting and model-selection history could not be fully reconstructed from the archive. It is therefore reported as an exploratory comparator with privileged inputs and cannot serve as a fair benchmark for image-only grading.

\subsubsection{Exploratory Architecture, Loss, and Proposal Comparisons}
\label{sec:ablations}

The experiment archive contains annotation-assisted configurations that vary the global backbone, gate, or loss weights, alongside automatic runs with different detector thresholds, detector families, or candidate limits. Table~\ref{tab:ablation_results} includes only configurations with a corresponding archived confusion matrix. The runs were not consistently matched for architecture, auxiliary-loss activity, annotation source, or random seed. Differences in their scores therefore cannot be attributed solely to the removal of an individual component.

\begin{table}[t]
\centering\footnotesize
\caption{Retrospective configuration comparisons on ACNE04 ($N=292$). All scores were recomputed from the corresponding archived confusion matrices. Oracle rows use supplied lesion boxes and differ in backbone or loss settings. The archive does not establish that detector operating-point comparisons were prespecified model-selection experiments conducted independently of the test set.}
\label{tab:ablation_results}
\renewcommand{\arraystretch}{1.09}
\begin{tabularx}{\columnwidth}{@{}Xrrr@{}}
\toprule
Configuration & Acc. (\%) & Macro-F1 & QWK \\
\midrule
\multicolumn{4}{@{}l}{\textit{Oracle inputs: distinct experiment configurations}} \\
ViT-only oracle, seed 42 (reference) & 96.23 & 0.9552 & 0.9773 \\ 
No gate (other backbone/loss settings) & 89.38 & 0.8415 & 0.9303 \\ 
CNN-only, no ViT (other settings) & 95.21 & 0.9417 & 0.9716 \\ 
No count-consistency (dual-trunk variant) & 94.18 & 0.9121 & 0.9653 \\ 
ViT-only, no count-consistency, seed 42 & 96.23 & 0.9526 & 0.9773 \\ 
No global-anchor KL (dual-trunk variant) & 95.55 & 0.9426 & 0.9735 \\ 
NO\_ORD script (dual-trunk variant) & 95.55 & 0.9426 & 0.9735 \\ 
NO\_RESREG script (dual-trunk variant) & 95.55 & 0.9487 & 0.9737 \\ 
\midrule
\multicolumn{4}{@{}l}{\textit{Automatic inputs: retrospective proposal operating points}} \\
YOLO11s, $\tau=0.050$, $K=16$ & 80.14 & 0.7875 & 0.8704 \\ 
YOLO11s, $\tau=0.075$, $K=16$ (selected) & 80.82 & 0.7877 & 0.8807 \\ 
YOLO11s, $\tau=0.150$, $K=16$ & 76.71 & 0.7196 & 0.8325 \\ 
YOLO11s, $\tau=0.075$, $K=32$ & 79.79 & 0.7819 & 0.8771 \\ 
YOLO12s, $\tau=0.075$, $K=16$ & 78.42 & 0.7588 & 0.8664 \\ 
\bottomrule
\end{tabularx}
\end{table}

The gate-free oracle run achieved 89.38\% accuracy and a QWK of 0.9303, compared with 96.23\% and 0.9773, respectively, for the selected ViT-only seed-42 oracle. This difference cannot be attributed to gate removal alone because the archived configurations also differ in backbone composition, active regularization terms, or both. The selected ViT-only training script assigned zero weight to the ordinal and residual-norm auxiliary penalties. The results therefore do not support claims that removing either penalty independently changes performance. A ViT-only seed-42 variant with zero count-consistency weight also achieved 96.23\% accuracy. Its macro-F1 was 0.9526, compared with 0.9552 for the corresponding selected oracle. This single-run comparison does not establish that count consistency is unnecessary in other settings.

Among the archived fully automated runs, confidence thresholds of 0.050, 0.075, and 0.150 yielded accuracies of 80.14\%, 80.82\%, and 76.71\%, respectively. At a threshold of 0.075, increasing the number of retained boxes from 16 to 32 yielded 79.79\% accuracy, while substituting the archived YOLO12s model yielded 78.42\%. Missed proposals, false candidates, candidate selection, and model-specific behavior may contribute to these differences, but their effects were not isolated. The run at a threshold of 0.100 and several previously listed ablations could not be matched unambiguously to saved confusion matrices. They are therefore excluded from the verified table pending confirmation of their provenance. This retrospective sweep does not establish unbiased selection of the detector threshold independently of the test results.

\subsubsection{ACNE04 Exact-Image Split-Integrity Findings}
\label{sec:data_hygiene}

SHA-256 comparisons identified 15 exact-image duplicate pairs across partitions in the executed ACNE04 fold: 10 train--test, three train--validation, and two validation--test pairs. Five pairs had conflicting folder-grade labels, comprising one Mild--Moderate pair and four Moderate--Severe pairs. Within-fold repeats were analysed separately and are excluded from these 15 cross-split pairs. The overlaps and label disagreements complicate the interpretation of internal benchmark results. Their precise effect on accuracy was not established, and they do not define a mathematical accuracy ceiling. Hash equality identifies byte-identical files only; it cannot rule out near-duplicate images or repeated participants across partitions.

\subsection{Frozen External Evaluation and Three-Dermatologist Annotation Sensitivity}
\label{sec:external_validation}

\subsubsection{External Cohort and Eligibility Findings}
\label{sec:external_cohort}

The archived PLSBRACNE01 evaluation included 200 subjects, each represented by frontal, left, and right facial photographs, for a total of 600 views. The recorded subject counts for the four Pillsbury grade indices were 58, 47, 65, and 30. Both the image collection and grading criteria differed from those used in the internal ACNE04 evaluation. Encoding the two grading systems as four ascending class indices does not establish clinical equivalence. Previous criterion-adaptation research describes PLSBRACNE01 as using a distinct grading system~\citep{lensrwLinKIEGLFN2022}.

The KIEGLFN study describes PLSBRACNE01~\citep{lensrwLinKIEGLFN2022} and attributes its collection to collaboration with the Dermatology Department of Heilongjiang Provincial Hospital. According to that publication, dermatologists from the same hospital and the Fourth Hospital of Harbin Medical University assisted with lesion annotation. The study also reports approval from the Institutional Ethics Committee of Heilongjiang Provincial Hospital.

The PLSBRACNE01 dataset used in this study was provided directly by Prof. Lin through a dataset access link shared for research use. The external evaluation comprised 200 participants, each represented by frontal, left lateral, and right lateral facial photographs, yielding 600 views in total. The dataset and its associated clinical grading framework
were originally described by
\citet{lensrwLinKIEGLFN2022}.

The archived hash audit found no exact-file duplicate views within the external evaluation set and no byte-identical overlap with the archived ACNE04 images. These checks alone do not establish patient independence.

\subsubsection{Crop-to-View Registration Yield}
\label{sec:registration}

The dermatologists' crop-level lesion boxes were mapped to the corresponding full facial views using SIFT--RANSAC homographies and the prespecified geometric acceptance criteria described in Section~\ref{sec:external_protocol}. Of the 450 eligible crop views with unambiguous source associations, 449 (99.78\%) met the recorded registration criteria. One right-view registration was rejected. Complete registration across all three views was available for 149 subjects. Requiring usable annotations from all three dermatologists yielded a common analysis cohort of 148 subjects and 444 photographic views. Inclusion in this subset depended on registration success and annotation availability; it was not a random sample of the full 200-subject cohort.

\subsubsection{Full-Cohort Automated Results and Post-hoc Failure Diagnostics}
\label{sec:automated_external}

In the frozen evaluation of all 200 participants, the independently trained ConvNeXtV2-L global classifier achieved 42.50\% accuracy, a macro-F1 of 0.4118, QWK of 0.5052, and MAE of 0.7050. Fully automatic LENS-GRF achieved 35.00\% accuracy, a macro-F1 of 0.3121, QWK of 0.3653, and MAE of 0.7600. The automatic-minus-global accuracy difference was $-7.50$ percentage points, with an unadjusted subject-level bootstrap 95\% confidence interval of $-14.00$ to $-0.50$ percentage points (not adjusted for multiple comparisons). Among discordant outcomes, 17 subjects were correctly classified only by automatic LENS-GRF and 32 only by the global classifier. The exact two-sided McNemar test yielded an unadjusted $p$-value of $0.0444$. After Holm adjustment for the archived family of two full-cohort comparisons, the $p$-value was $0.0888$, which did not meet the family-wise significance threshold of $0.05$.

Automatic LENS-GRF achieved 50.52\% macro precision, 34.31\% macro sensitivity (balanced accuracy), and 78.19\% macro one-versus-rest specificity on the full cohort. Post-hoc analysis of the retained detector traces showed a mean of 17.24 raw candidate boxes per view and a maximum of 96. Of the 600 views, 247 (41.17\%) exceeded the 16-box retention limit. Mean gate activation was 0.8815, with a per-view standard deviation of 0.0855. After averaging gate activation across each subject's three views, the across-subject standard deviation was 0.0699. Subjects with reference Grade~4 had a mean subject-level gate activation of 0.9221. High gate activation occurred alongside poor external accuracy, and box count was positively associated with gate activation. These observations do not establish detector errors or gate behavior as the sole causes of the performance loss. Image framing, apparent lesion scale, and grading criteria changed simultaneously, and their respective effects could not be separated in this analysis.

\begin{figure*}[t]
\centering
\begin{subfigure}[b]{0.48\textwidth}
\centering\includegraphics[width=\textwidth]{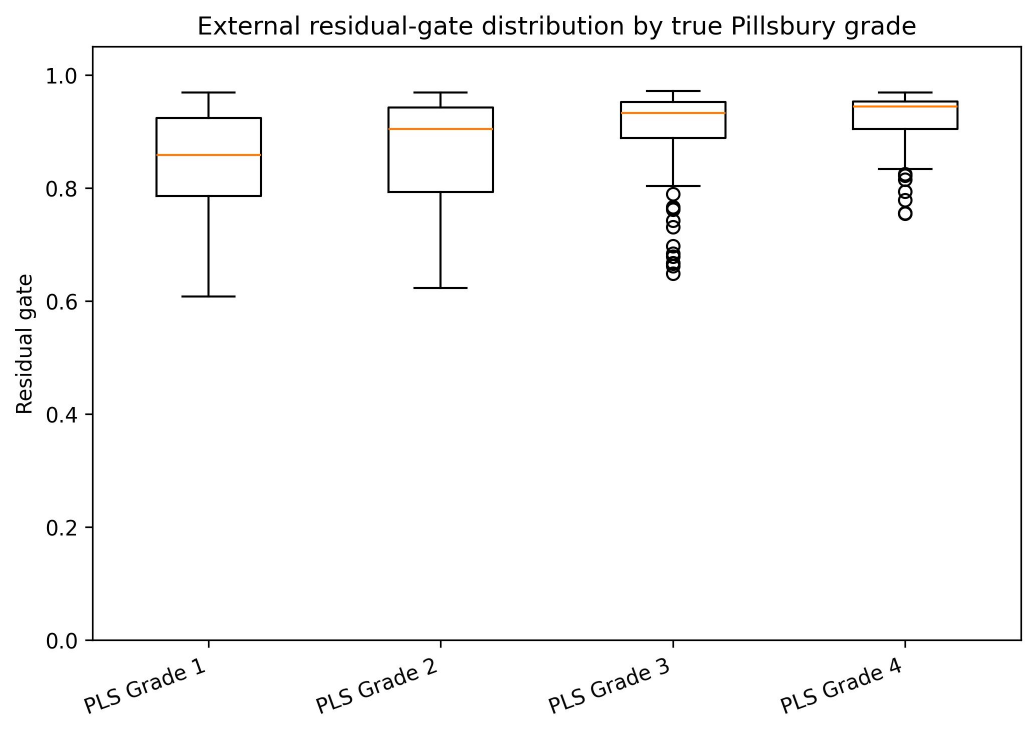}
\caption{Gate distribution by reference Pillsbury grade.}
\label{fig:gate_grade}
\end{subfigure}\hfill
\begin{subfigure}[b]{0.48\textwidth}
\centering\includegraphics[width=\textwidth]{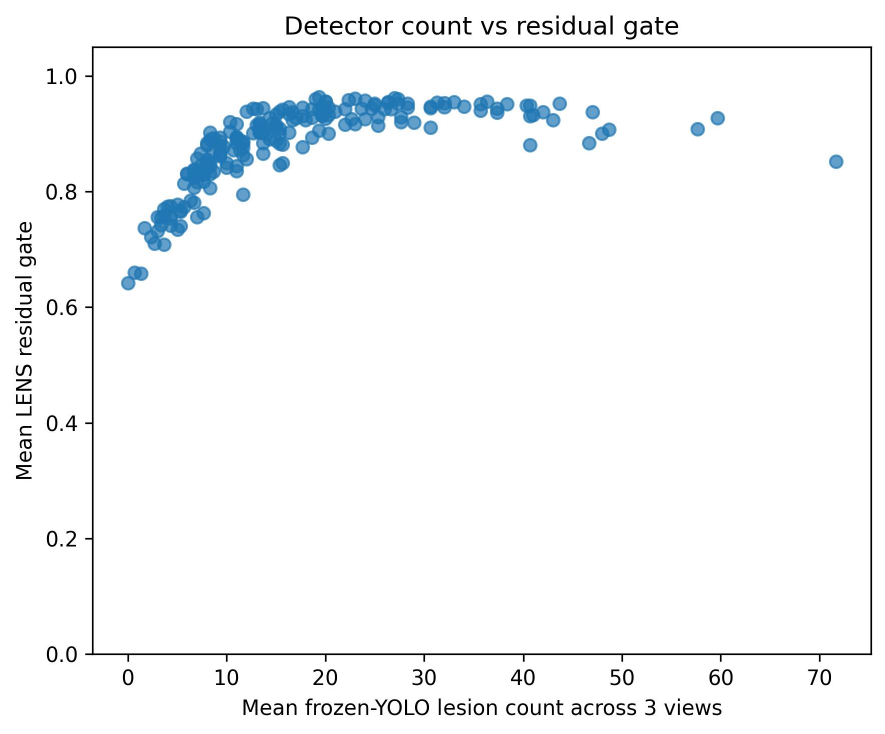}
\caption{Raw proposal count and the residual gate.}
\label{fig:count_gate}
\end{subfigure}
\caption{Post-hoc gate diagnostics for frozen automated PLSBRACNE01 evaluation. The view-level gate mean was 0.8815; apparent count--gate association is descriptive, not a causal mechanism test.}
\label{fig:failure_diagnostics}
\end{figure*}

\subsubsection{Matched Annotation-Assisted Oracle Evaluation}
\label{sec:oracle_results}

Each annotation-assisted oracle configuration and the automatic comparator were evaluated on the same 148 subjects (444 views). Table~\ref{tab:external_metrics} reports subject-level performance and confidence intervals obtained by bootstrapping subjects. Table~\ref{tab:paired_comparisons} reports exact paired tests with Holm adjustment across the specified family of nine comparisons. These matched-cohort estimates are separate from those for the full 200-subject cohort.

\begin{table*}[t]
\centering\small
\caption{Matched PLSBRACNE01 cohort ($N=148$ subjects; 444 views). Brackets give 95\% confidence intervals from 5,000 subject-level bootstrap resamples for the displayed metrics. Boxes supplied by the doctors are privileged evaluation inputs.}
\label{tab:external_metrics}
\renewcommand{\arraystretch}{1.08}
\resizebox{\textwidth}{!}{%
\begin{tabular}{@{}lcccccc@{}}
\toprule
Model & Accuracy (\%) & B-Acc. (\%) & Macro-F1 & QWK & Grade MAE & Brier \\
\midrule
Global ConvNeXtV2-L & 41.22 [33.1,49.3] & 42.54 & 0.4091 [0.326,0.485] & 0.5033 [0.372,0.620] & 0.7230 [0.608,0.838] & 0.8624 \\
Automatic LENS-GRF & 35.14 [27.7,43.2] & 34.81 & 0.3124 [0.234,0.391] & 0.3861 [0.270,0.497] & 0.7635 [0.655,0.865] & 0.9377 \\
LENS internal ViT & 31.08 [23.6,38.5] & 30.73 & 0.2622 [0.190,0.337] & 0.2145 [0.106,0.329] & 1.0068 [0.865,1.142] & 0.7431 \\
\midrule
GT oracle: Doctor A & 47.97 [39.8,56.1] & 47.08 & 0.4595 [0.364,0.543] & 0.5620 [0.450,0.657] & 0.5878 [0.493,0.689] & 0.7254 \\
GT oracle: Doctor B & 43.24 [35.8,51.4] & 41.52 & 0.4020 [0.316,0.484] & 0.5799 [0.485,0.661] & 0.6014 [0.514,0.689] & 0.7821 \\
GT oracle: Doctor C & 44.59 [36.5,52.7] & 44.16 & 0.4035 [0.327,0.478] & 0.5340 [0.431,0.625] & 0.6284 [0.527,0.730] & 0.7857 \\
\bottomrule
\end{tabular}}
\end{table*}

\begin{table*}[t]
\centering\small
\caption{Nine exact two-sided McNemar tests for paired correctness on the matched 148 subjects. Holm values control the family-wise error rate across the displayed nine comparisons. Delta denotes first model minus second model. Confidence intervals were computed by subject-level bootstrap.}
\label{tab:paired_comparisons}
\renewcommand{\arraystretch}{1.08}
\begin{tabularx}{\textwidth}{@{}Xlrrrrr@{}}
\toprule
Pair (first vs. second) & Discordant (first/second) & $\Delta$Acc. (pp) & 95\% CI (pp) & $\Delta$QWK & Exact $p$ & Holm $p$ \\
\midrule
Doctor A oracle vs. automatic LENS-GRF & 34/15 & +12.84 & [+4.04,+22.30] & +0.1758 & 0.0094 & 0.0846 \\
Doctor B oracle vs. automatic LENS-GRF & 34/22 & +8.11 & [-1.35,+18.24] & +0.1938 & 0.1409 & 0.9863 \\
Doctor C oracle vs. automatic LENS-GRF & 31/17 & +9.46 & [+0.68,+18.24] & +0.1478 & 0.0595 & 0.4757 \\
\midrule
Doctor A oracle vs. independent global & 36/26 & +6.76 & [-4.05,+17.57] & +0.0586 & 0.2529 & 1.0000 \\
Doctor B oracle vs. independent global & 38/35 & +2.03 & [-9.46,+13.51] & +0.0765 & 0.8151 & 1.0000 \\
Doctor C oracle vs. independent global & 28/23 & +3.38 & [-6.08,+12.84] & +0.0306 & 0.5758 & 1.0000 \\
\midrule
Doctor A oracle vs. Doctor B oracle & 28/21 & +4.73 & [-4.05,+13.51] & -0.0179 & 0.3916 & 1.0000 \\
Doctor A oracle vs. Doctor C oracle & 25/20 & +3.38 & [-6.08,+12.16] & +0.0280 & 0.5515 & 1.0000 \\
Doctor B oracle vs. Doctor C oracle & 31/33 & -1.35 & [-12.16,+9.46] & +0.0459 & 0.9007 & 1.0000 \\
\bottomrule
\end{tabularx}
\end{table*}

Using Doctor~A's lesion annotations, LENS-GRF achieved 47.97\% accuracy on the matched cohort, compared with 35.14\% for automatic inference, a difference of 12.84 percentage points. The McNemar test yielded an unadjusted $p$-value of $0.0094$ and a Holm-adjusted $p$-value of $0.0846$; the comparison therefore did not meet the family-wise significance threshold of $0.05$. Doctor~B's annotations yielded the highest oracle QWK ($0.5799$), while Doctor~C's annotations yielded 44.59\% accuracy. The oracle configuration using Doctor~A's annotations also exceeded the independent global comparator by 6.76 percentage points, with a paired exact $p$-value of $0.2529$ (Holm-adjusted $p=1.0000$). These results show sensitivity to the source of lesion annotations but do not establish an independently validated improvement in clinical effectiveness.

\begin{table}[t]
\centering\scriptsize
\caption{Grade-wise precision (P), recall (R), and F1 for matched 148 subjects. Counts by grade are 39, 34, 50, and 25.}
\label{tab:per_class_metrics}
\renewcommand{\arraystretch}{1.06}
\resizebox{\columnwidth}{!}{%
\begin{tabular}{lcccc}
\toprule
Model / metric & Grade 1 & Grade 2 & Grade 3 & Grade 4 \\
\midrule
Independent global: P/R & 0.688/0.564 & 0.318/0.618 & 0.357/0.200 & 0.364/0.320 \\
Independent global: F1 & 0.6197 & 0.4200 & 0.2564 & 0.3404 \\
\midrule
Automatic LENS-GRF: P/R & 0.571/0.308 & 0.274/0.765 & 0.414/0.240 & 0.667/0.080 \\
Automatic LENS-GRF: F1 & 0.4000 & 0.4031 & 0.3038 & 0.1429 \\
\midrule
Doctor A oracle: P/R & 0.800/0.410 & 0.349/0.853 & 0.525/0.420 & 1.000/0.200 \\
Doctor A oracle: F1 & 0.5424 & 0.4957 & 0.4667 & 0.3333 \\
\midrule
Doctor B oracle: P/R & 0.857/0.154 & 0.306/0.647 & 0.492/0.580 & 0.700/0.280 \\
Doctor B oracle: F1 & 0.2609 & 0.4151 & 0.5321 & 0.4000 \\
\midrule
Doctor C oracle: P/R & 0.880/0.564 & 0.330/0.882 & 0.414/0.240 & 0.667/0.080 \\
Doctor C oracle: F1 & 0.6875 & 0.4800 & 0.3038 & 0.1429 \\
\bottomrule
\end{tabular}}
\end{table}

\begin{figure*}[t]
\centering
\begin{subfigure}[b]{0.24\textwidth}
\centering\includegraphics[width=\textwidth]{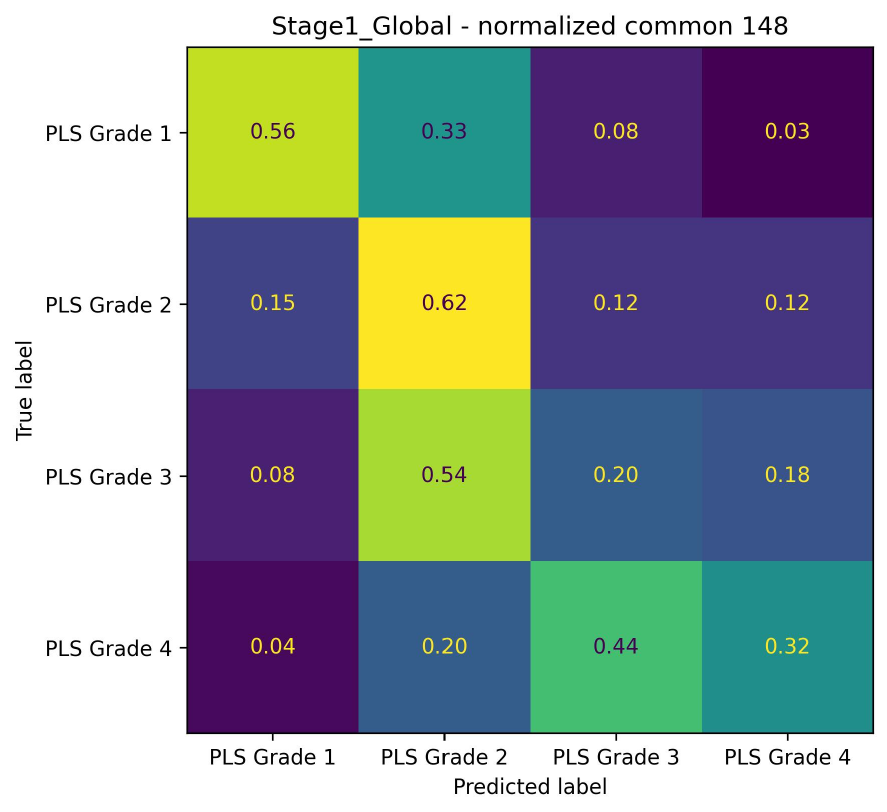}
\caption{Independent global}\label{fig:cm_stage1}
\end{subfigure}\hfill
\begin{subfigure}[b]{0.24\textwidth}
\centering\includegraphics[width=\textwidth]{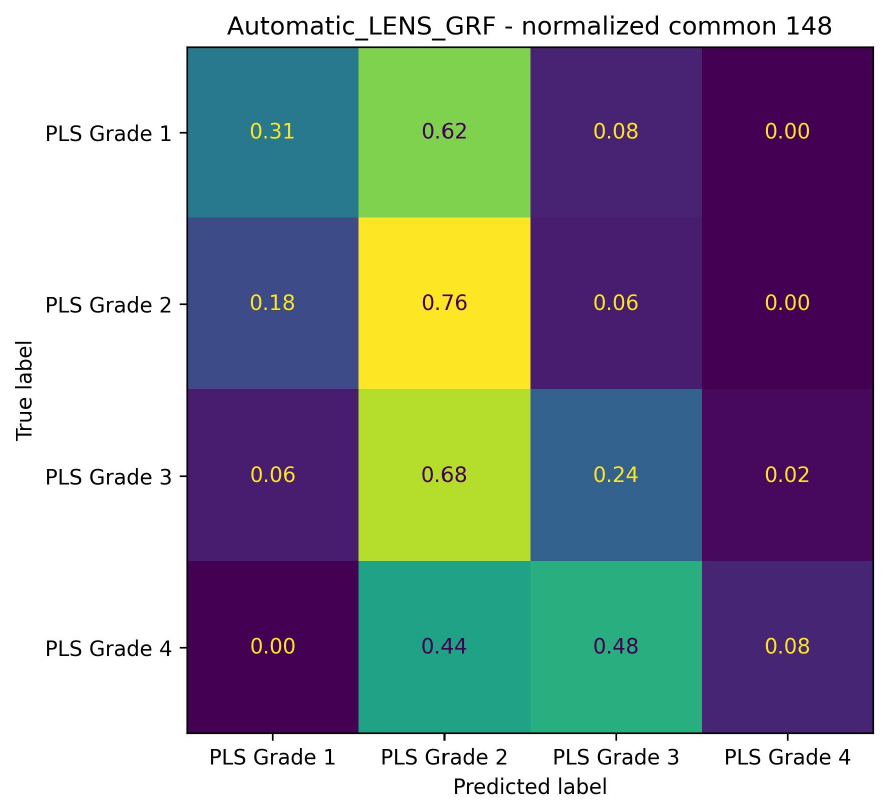}
\caption{Automatic LENS-GRF}\label{fig:cm_auto}
\end{subfigure}\hfill
\begin{subfigure}[b]{0.24\textwidth}
\centering\includegraphics[width=\textwidth]{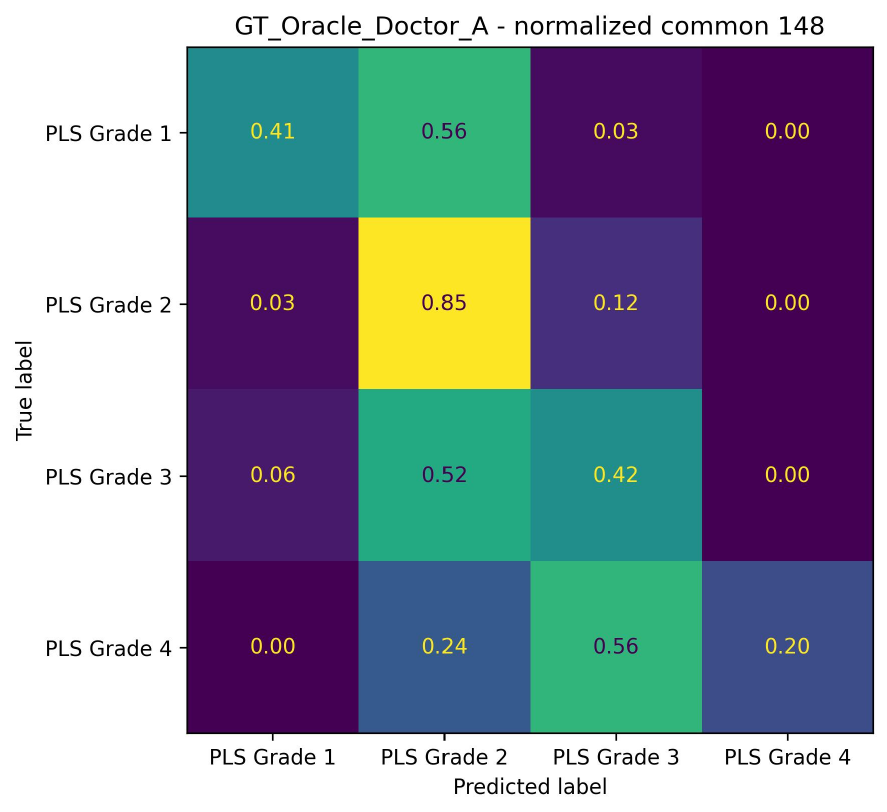}
\caption{Doctor A boxes}\label{fig:cm_doc_a}
\end{subfigure}\hfill
\begin{subfigure}[b]{0.24\textwidth}
\centering\includegraphics[width=\textwidth]{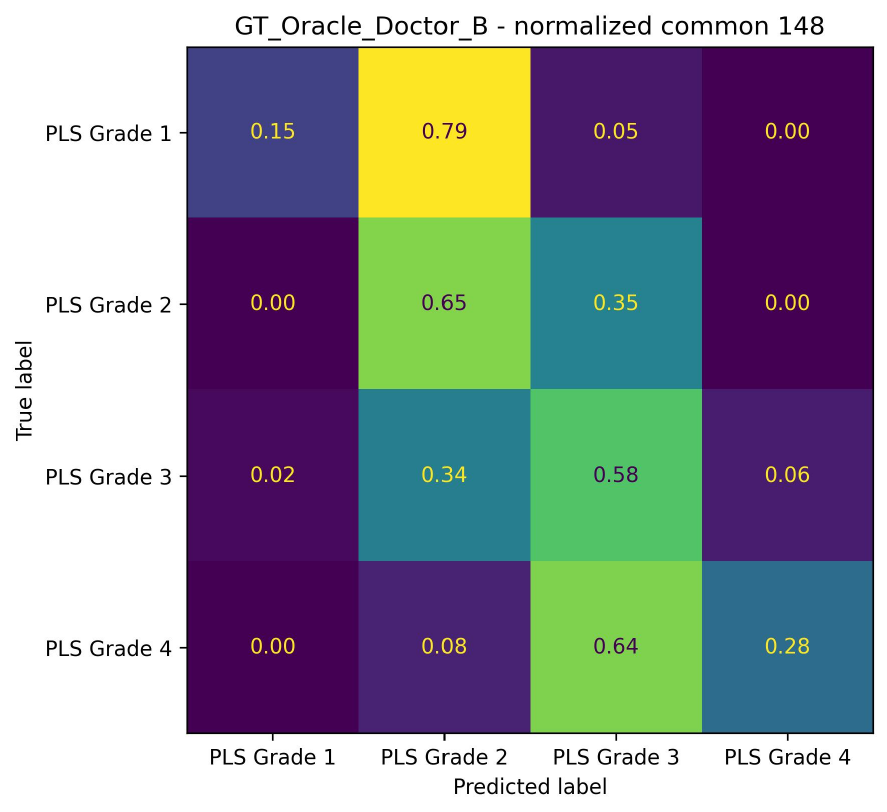}
\caption{Doctor B boxes}\label{fig:cm_doc_b}
\end{subfigure}
\caption{Normalized subject-level confusion matrices on the common 148-subject external cohort. The panels illustrate changes in the distribution of predictions under alternative input evidence, not direct dermatologist severity agreement.}
\label{fig:confusion_matrices}
\end{figure*}

\subsubsection{Grade-Specific Performance and Annotation-Conditioned Agreement}
\label{sec:inter_rater}

In the matched cohort, automatic LENS-GRF correctly identified two of the 25 subjects with Pillsbury Grade~4, giving a sensitivity of 0.08. Sensitivity was 0.20 with Doctor~A's lesion boxes and 0.28 with Doctor~B's boxes. The low automatic sensitivity limits confidence in severe-grade recognition under the evaluated conditions. Oracle predictions based on different dermatologists' boxes agreed for 49.32\%--66.22\% of subjects (Table~\ref{tab:inter_rater_agreement}). These comparisons measure \emph{agreement between model predictions conditioned on different lesion annotations}. They do not measure agreement among clinicians' independent severity assessments or define a human-performance ceiling.

\begin{table}[t]
\centering\footnotesize
\caption{Agreement between annotation-conditioned oracle predictions on 148 matched subjects. Lesion-count summaries are the mean \emph{per photographic view}, not the total number per participant; medians are medians of each participant's three-view mean.}
\label{tab:inter_rater_agreement}
\renewcommand{\arraystretch}{1.07}
\resizebox{\columnwidth}{!}{%
\begin{tabular}{lrrrr}
\toprule
Oracle pair & Agreement (\%) & QWK & Mean boxes/view & Median subj. mean/view \\
\midrule
A vs. B & 60.14 & 0.6046 & 15.47 vs. 21.86 & 12.83 vs. 17.83 \\
A vs. C & 66.22 & 0.6478 & 15.47 vs. 12.96 & 12.83 vs. 10.50 \\
B vs. C & 49.32 & 0.4823 & 21.86 vs. 12.96 & 17.83 vs. 10.50 \\
\bottomrule
\end{tabular}}
\end{table}

Doctor~B's mean annotation count was 21.86 per view, approximately 68.6\% higher than Doctor~C's mean of 12.96. The medians of participants' mean counts across views also differed (Table~\ref{tab:inter_rater_agreement}). Applying Hayashi thresholds to annotation counts alone yielded accuracies of 44.59\%, 44.59\%, and 41.22\% for Doctors~A, B, and C, respectively, on the matched cohort. These counts include all annotated categories and photographic views and were not verified as unilateral inflammatory-lesion counts. The mapping therefore provides an exploratory assessment of mismatch between grading criteria and should not be treated as a clinical reference grade. The three-rater comparison is limited to the registered subset of 148 subjects and cannot replace full-cohort automatic validation.





\section{Discussion}
\label{sec:discussion}

\subsection{Principal Findings Across Internal and External Evaluation}
\label{sec:discussion_principal}

The internal and external evaluations show that the benefit of lesion-guided acne grading depends on the source of lesion evidence and the evaluation setting. On the ACNE04 test partition, fully automatic LENS-GRF achieved 80.82\% accuracy, compared with 78.42\% for the ViT-based global comparator and 77.74\% for the separately trained ConvNeXtV2-L baseline. These differences suggest that selected detector proposals can complement whole-image information within the development benchmark, although the two global baselines are distinct comparators. With ground-truth lesion boxes, the annotation-assisted model achieved a mean accuracy of $95.89\%\pm0.59\%$ across three seeds. The gap between automatic and annotation-assisted performance shows sensitivity to the source of local evidence. The oracle result does not represent autonomous detector-based performance or a theoretical maximum for the grading task.

The retrospective internal comparisons included configurations that differed in gate use, global backbones, auxiliary objectives, and lesion-proposal settings. Several runs also differed in other architectural settings, active regularization terms, or random seeds. These differences prevent causal attribution of changes in performance to the removal of any single component.  A recorded ViT-only seed-42 run without count consistency matched the selected ViT-only oracle's accuracy of 96.23\%. The previously reported direct-fusion comparison could not be independently linked to an archived confusion matrix. These exploratory comparisons do not isolate individual component effects or establish gate reliability under external distribution shift.

The frozen PLSBRACNE01 evaluation showed a different performance pattern. Across 200 subjects and 600 photographic views, automatic LENS-GRF achieved 35.00\% subject-level accuracy, below the independent ConvNeXtV2-L baseline at 42.50\%. PLSBRACNE01 differs from ACNE04 in image framing and apparent lesion scale, and its reference categories follow Pillsbury rather than Hayashi-oriented grading. Image characteristics and clinical targets therefore changed together. Without a controlled intervention, the contribution of each change to the performance decline cannot be isolated. The external result limits claims that the internal improvement will generalize to other datasets.

On the matched cohort of 148 subjects with valid annotation registrations from all three dermatologists, accuracy increased from 35.14\% with automatic proposals to 47.97\% with Doctor~A's lesion boxes. Doctor~B's boxes produced the highest oracle quadratic weighted kappa, at 0.5799. The Doctor~A comparison yielded an unadjusted exact McNemar $p$-value of 0.0094, but the Holm-adjusted value was 0.0846 and did not meet the family-wise significance threshold of 0.05. Automatic Grade~4 sensitivity was also only 0.08 in this cohort. These findings show that model performance varies with the supplied annotations, but they do not establish clinical benefit or superiority after correction for multiple comparisons.

\subsection{Interpretation in Relation to Prior Acne-Grading Research}
\label{sec:discussion_prior}

Count-aware modeling is established in acne severity assessment. \citet{lensrwWu2019} jointly modeled severity and lesion counts using label distributions on ACNE04, while \citet{lensintroProkhorov2024} investigated smoothing across adjacent severity categories. These approaches address uncertainty within a given grading system. In the present external evaluation, the grading system itself also changes. \citet{lensintroChen2025} examined feature-feedback pseudo-label learning across ACNE04 and the Bernardis-based ACNE-ECKH dataset, explicitly addressing learning across clinical standards. However, their datasets and training protocol differ from the frozen Hayashi-to-Pillsbury transfer evaluated here.

Global--local fusion also has established precedents in acne grading. \citet{lensrwLinPrior2022} used lesion-related prior information to guide image-based severity prediction. Their subsequent KIEGLFN framework combined key-information enhancement with global--local representations and examined fine-tuning on a smaller Pillsbury-based cohort after training on ACNE04~\citep{lensrwLinKIEGLFN2022}. Diagnostic Evidence Distillation transferred lesion-related knowledge from a teacher to an image-based student, reducing the need for the same explicit local annotations at inference~\citep{lensrwLinDED2023}. These methods handle evidence availability and adaptation between grading criteria differently from the detector-dependent residual refinement used in LENS-GRF. Comparing their reported accuracy requires accounting for differences in training exposure, lesion inputs, clinical labels, and evaluation units.

Lesion-detection studies address a different stage of the grading pipeline. ACNet was designed to handle variation in illumination, lesion size, and lesion density~\citep{lensrwMin2021}. A more recent comparison of YOLO variants evaluated acne localization and count-derived severity on ACNE04~\citep{lensrwKhairani2025}. These findings inform candidate generation, but they do not establish how a frozen classifier performs when predictions from multiple facial views are combined at the subject level. AcneAI combines lesion segmentation and per-lesion scoring to produce a continuous image-level severity measure~\citep{lensrwGazeauAcneAI}. Its output scale and evaluation tasks also differ from the four-class external endpoint used here.

Experiments with SimCLR-based acne representations found that supervised transfer learning outperformed generic contrastive pretraining under the evaluated conditions~\citep{lensintroSrijiranon2026}. A systematic review also identified limited independent and prospective validation in acne-AI research~\citep{lensintroTraini2025}. These findings support evaluating generalization separately from internal performance. The present frozen-transfer experiment documents a failure under a defined external protocol; it does not invalidate earlier methods or establish a new performance ranking.

\subsection{Implications for Lesion-Evidence Fusion and Dermatological AI}
\label{sec:discussion_implications}

The residual formulation provides a conditional mathematical identity, but the reliability of the learned gate requires separate evaluation. Fused logits equal global logits when the scalar gate is zero. The implemented sigmoid gate, however, is not constrained to reach zero when lesion proposals are missing or incorrect. In the frozen external evaluation, the mean gate remained approximately 0.8815 despite the decline in automatic grading accuracy. The identity condition therefore does not provide an operational failure detector, a calibrated uncertainty estimate, or a clinical safety guarantee. Whether a separately calibrated rejection mechanism or an explicit fallback rule would improve performance remains untested.

The diagnostic traces suggest that the detector may limit the quality of local evidence. External views contained a mean of 17.24 detector boxes, and 41.17\% exceeded the 16-token retention limit. False candidates and truncation may distort the lesion representation, particularly in images with extensive inflammatory disease. The low Grade 4 sensitivity makes this possibility clinically relevant. However, the observed box counts and gate behavior are post hoc findings and do not establish which mechanisms caused the errors. The concurrent shift from count-oriented Hayashi classes to Pillsbury grades, which also reflect morphology and extent, remains another possible source of disagreement.

The three dermatologists' annotation sets produced different model predictions. Pairwise agreement between the resulting oracle predictions ranged from 49.32\% to 66.22\%. These values measure variation in model outputs conditioned on different lesion annotations; they do not measure agreement among the dermatologists' severity diagnoses or define a human-performance ceiling. The matched analysis also included only the 148 subjects with valid registrations for all three annotation sets. Its estimates therefore cannot replace those from the full 200-subject automatic evaluation. These findings support reporting annotation sources, inclusion criteria, grade-specific errors, and uncertainty when assessing lesion-assisted grading systems.

Exact-image duplicates and conflicting labels across ACNE04 partitions further complicate the interpretation of internal performance. Hash comparisons identify exact file overlap but cannot establish patient independence or exclude visually similar duplicates. Future studies could test scale-aware proposal generation, calibrated candidate selection, targets that account for differences between grading criteria, and selective residual fallback using prespecified protocols. The present study has not shown that any of these changes improves frozen external grading. Clinical use would require independent validation against clearly defined reference standards.




\subsection{Limitations}
\label{sec:limitations}

In this retrospective study, an exact-hash audit of the archived ACNE04 split identified duplicate images across partitions, including images assigned different severity grades. This overlap complicates the interpretation of internal performance, although its effect was not quantified and patient-level independence was not established. Exact hashes cannot identify recropped images or photographs of the same patient when the file contents differ. We did not evaluate an alternative split with separate patient groups in each partition. Retrospective detector-threshold sweeps also do not demonstrate that the final operating point was selected using validation data alone. Comparisons among these settings should therefore be considered exploratory.

External evaluation involved simultaneous changes in image framing, apparent lesion scale, and severity criteria. ACNE04 uses Hayashi-oriented grades that emphasize lesion counts, whereas PLSBRACNE01 uses Pillsbury-oriented categories that also account for lesion morphology and distribution. Encoding both scales as four ordered classes does not make them clinically equivalent. Because imaging conditions and reference criteria changed together, we could not isolate their respective effects on external performance. The three-dermatologist oracle analysis included only the 148 participants with complete valid registrations from the full cohort of 200. Results from this selected subset cannot be extended to excluded participants without further evidence.

The lesion encoder retained at most $K_{\max}=16$ selected proposals per view, although 247 of the 600 external views ($41.17\%$) produced more candidate boxes. We did not establish whether the discarded proposals contained true lesions or how much truncation contributed to grading errors. The sigmoid gate was not explicitly clamped to zero when detections were absent or unreliable. The mathematical identity at a zero gate therefore does not provide an implemented safeguard against detection failure. Component ablations also differed in model settings, random seeds, or the availability of lesion annotations, limiting the extent to which external performance could be attributed to individual modules.

The supplied dermatologist boxes were privileged inputs for oracle evaluation and were unavailable to the autonomous grader. Agreement between model predictions generated from these annotation sets does not measure agreement among clinicians' severity diagnoses or define a human-performance ceiling. Automatic sensitivity for Pillsbury Grade~4 was $8\%$ (2 of 25 participants) in the matched external cohort, limiting confidence in severe-grade recognition. External predictions were obtained by arithmetically averaging predictions across three facial views; a learned model for combining views was not evaluated. Prospective clinical utility, performance across demographic subgroups, and robustness to unconstrained point-of-care imaging remain unestablished.




\section{Conclusions and Future Work}
\label{sec:conclusion}

We investigated whether permutation-invariant aggregation of selected lesion-region tokens could refine global Vision Transformer severity logits through the gated residual update in LENS-GRF. On ACNE04, the fully automatic pipeline achieved $80.82\%$ test accuracy. A separately evaluated annotation-assisted oracle configuration achieved a mean accuracy of $95.89\% \pm 0.59\%$ across three seeds using ground-truth lesion boxes. The oracle configuration had access to annotations unavailable to the automatic pipeline, so its accuracy does not represent autonomous grading performance or a theoretical accuracy ceiling. Split-integrity issues also limit the interpretation of the internal results.

In the frozen evaluation on the full cohort of 200 PLSBRACNE01 participants, automatic LENS-GRF achieved $35.00\%$ subject-level accuracy, compared with $42.50\%$ for an independently trained ConvNeXtV2-L global baseline. In the separate matched cohort of 148 participants, automatic accuracy was $35.14\%$, while accuracy using Doctor~A's lesion annotations was $47.97\%$. The paired comparison yielded an unadjusted exact McNemar $p$-value of $0.0094$, but did not meet the $0.05$ significance threshold after Holm adjustment ($p=0.0846$). The oracle configuration using Doctor~B's annotations achieved the highest quadratic weighted kappa ($0.5799$). The dependence on supplied annotations and the low severe-grade sensitivity limit conclusions about automated generalization and clinical effectiveness.

The residual formulation returns the global logits when the gate is zero, but the study did not demonstrate that the learned gate automatically suppresses unreliable detections. Future work should test scale-aware lesion proposals, candidate retention strategies, independently selected detector operating points, and calibrated or explicitly enforced selective fallback under distribution shift. Targets that account for differences in clinical grading criteria and learned aggregation across facial views also require controlled comparisons with the current approach. Independent prospective evaluations should use clearly defined clinical reference standards and assess severe-grade sensitivity, calibration, and performance across patient subgroups and image acquisition conditions. The benefits of these proposed changes remain untested, and the present evidence does not support clinical deployment.


\section*{Acknowledgements}
The authors acknowledge American International University-Bangladesh (AIUB) for its support and contribution to this work. The authors also acknowledge the Advanced Machine Intelligence Research Lab (AMIR) for its support and contribution.

\section*{CRediT Authorship Contribution Statement}
\textbf{Muhammad Muhtasim Shahriar}: Conceptualization, Methodology, Software, Validation, Formal analysis, Investigation, Data Curation, Writing - Original Draft, Writing - Review \& Editing, Visualization, Project administration.

\textbf{M.\,F. Mridha:} Supervision, Project Administration, Methodology, Resources, Writing -- Review \& Editing.

\textbf{Md. Rakibul Islam:} Supervision, Project Administration, Methodology, Resources, Funding Acquisition, Writing -- Review \& Editing.

\section*{Declaration of Competing Interest}
The authors declare that they have no known competing financial interests or personal relationships that could have appeared to influence the work reported in this paper.



\section*{Data and Code Availability}

The ACNE04 dataset was introduced by
\citet{lensrwWu2019}. The original dataset
and related research resources are described at
\url{https://github.com/xpwu95/LDL}.

The PLSBRACNE01 dataset was accessed through
a research-use link provided directly by
Prof. Lin, whose original study describes
the dataset~\citep{lensrwLinKIEGLFN2022}.
The clinical photographs are not redistributed
as part of this manuscript. Requests for
dataset access should be directed to the
original dataset custodians.

\section*{Declaration of Generative AI and
AI-Assisted Technologies in the Manuscript
Preparation Process}

All experimental results, numerical values, statistical
analyses, tables, and figures reported in this study were
produced directly from the authors' implemented code,
executed experiments, and verified experimental outputs.
No generative AI system was used to generate, modify,
fabricate, or infer experimental data, numerical results,
statistical outcomes, plots, figures, or scientific
conclusions. No AI-based image-generation system was
used to create the manuscript figures or the methodological
architecture diagram.

Generative AI tools were used only during manuscript
preparation for language refinement and organisational
assistance. The authors independently verified, revised,
and take full responsibility for the final scientific content
of the manuscript.


\input{references_numbered_RIE.tex}

\end{document}

%% file: sections/02_methods.tex

\section{Methods}
\label{sec:methodology}

\subsection{Study Design, Research Tasks, and Evaluation Units}
\label{sec:formulation}

We conducted a retrospective computational study of four-class acne severity grading using two previously released photographic datasets. Internal model development and independent cross-dataset evaluation were conducted separately. The proposed method, LENS-GRF, combines a global image representation with local lesion evidence derived from bounding boxes.

The study distinguished three evaluation regimes throughout design and reporting: (i) a global image-only classifier, (ii) fully automated fusion using detector-generated lesion proposals, and (iii) annotation-assisted (``oracle'') fusion using existing lesion boxes. The oracle boxes were research annotations and would not be available to an autonomous clinical pipeline. In the automated and oracle experiments, we also distinguished the global representation within LENS-GRF from the separately trained ConvNeXtV2 global comparator.

ACNE04 was used for model development, with each image providing one classification target and one evaluation observation. For external evaluation on PLSBRACNE01, the frozen ACNE04-trained systems processed three facial photographs per participant. Class probabilities were averaged across the frontal, left, and right views to obtain one subject-level prediction. The participant was therefore the primary external analysis unit. A matched subset with successfully registered lesion annotations from all three dermatologists was used to assess sensitivity to the annotation source. The external evaluation examined transfer across image-acquisition conditions and severity-label criteria; it was not a prospective clinical deployment study.

\subsection{Datasets, Severity-Label Systems, Cohorts, and Partitions}
\label{sec:datasets}

ACNE04 contained 1,457 photographs with image-level severity grades and lesion bounding-box files. The archived experimental fold comprised 1,049 training images, 116 validation images, and 292 test images. The audit of folder labels identified 497 Mild, 637 Moderate, 186 Severe, and 137 Very Severe images. The corresponding training/validation/test counts were 365/33/99 for Mild, 450/56/131 for Moderate, 138/15/33 for Severe, and 96/12/29 for Very Severe. These counts describe the local fold used in this study and may differ from class distributions reported for other ACNE04 partitions.

The Hayashi-related grading framework considers inflammatory lesion burden within a defined facial region. Counts of all boxes in an annotation or detector file therefore cannot be assumed to reproduce the required unilateral inflammatory-lesion count.

PLSBRACNE01 contained 200 participants, each represented by one frontal and two lateral photographs, giving 600 images. The four subject-level Pillsbury grades contained 58, 47, 65, and 30 participants, respectively. The registered subset common to all three raters comprised 148 participants and 444 photographs. Its inclusion criteria are specified below.

For computational transfer, the Hayashi-oriented Mild--Very Severe labels in ACNE04 and the Pillsbury Grades~1--4 in PLSBRACNE01 were encoded in ascending four-class order. Matching class indices do not establish clinical equivalence between the two scales. Table~\ref{tab:methods_datasets} summarizes the cohort sizes, evaluation units, and sources of lesion evidence.

\begin{table*}[t]
\centering
\caption{Study datasets, clinical criteria, analysis units and lesion-annotation availability. External photographs are nested within participants; the three-rater subset is contained in the full external cohort.}
\label{tab:methods_datasets}
\small
\begin{tabularx}{\textwidth}{@{}p{2.25cm}p{2.3cm}p{2.0cm}p{2.45cm}X@{}}
\toprule
Dataset / cohort & Clinical grade & Split / selection & Evaluation unit and size & Lesion-box availability and study role \\
\midrule
ACNE04 train & Hayashi-oriented, 4 classes & Training fold & 1,049 images & Annotation files available; automatic detector-proposal or annotation-box training by regime \\
ACNE04 validation & Same & Validation fold & 116 images & Annotation / proposal inputs matched by regime; checkpoint selection \\
ACNE04 test & Same & Held-out fold & 292 images & Internal image-level comparison and retrospective threshold diagnostics \\
PLSBRACNE01 full & Pillsbury, 4 grades & Independent full cohort & 200 subjects; 600 views & Frozen automatic inference; crop-level expert boxes not available for every subject \\
PLSBRACNE01 common & Same & Three-rater matched subset & 148 subjects; 444 views & Registered boxes from Doctors A, B and C; annotation-source sensitivity analysis \\
\bottomrule
\end{tabularx}
\end{table*}

\subsection{Dataset Integrity, Duplicate Identification, and Leakage Boundaries}
\label{sec:data_integrity}

We indexed the source images in the archived ACNE04 fold by partition, folder grade, associated lesion annotations, and SHA-256 file hash. The audit identified matching hashes across partitions, retained the source paths, and checked whether the corresponding folder grades agreed. The audit ledger recorded duplicate \emph{pairs} separately from unique image files and distinguished each pair's partition intersection from the full group of files sharing a hash.

The original fold assignments were preserved to maintain comparability with the recorded benchmark experiments. Overlapping files were not retrospectively removed, and test images were not relabeled. Audit counts and examples came from archived CSV records rather than filename similarity alone.

Matching SHA-256 hashes identify byte-identical files but do not establish whether patients are independent across partitions. Different hashes also do not exclude recropped images, photographs from the same acquisition sequence, visually similar images, or repeated participants. The audit therefore assesses exact-file overlap without establishing independence across all clinical image groups. The external source-image audit compared hashes both within the external collection and against the archived ACNE04 images. Uncertainty about patient overlap, conflicting labels, and differences in clinical criteria was retained as a limitation; these findings were not used to infer a deterministic accuracy ceiling.

\subsection{Facial Preprocessing and Resolution/Coordinate Handling}
\label{sec:afss}

The archived training code names the preprocessing class \texttt{AFSS\_YCrCb}. Its implementation generated a mask using fixed YCrCb thresholds: $Y\in[0,255]$, $Cr\in[133,173]$, and $Cb\in[77,127]$. The mask underwent one morphological opening and one dilation with a $6\times6$ elliptical structuring element, followed by Gaussian edge feathering with $\sigma=2$. The feathered mask was then used to alpha-blend the RGB image with a black background. We retain the historical AFSS name, although this implementation neither learned the thresholds nor adapted them to illumination.

In the archived automatic and oracle training scripts, preprocessing was applied to the loaded image before both the global-image transform and lesion ROI extraction. The global stream used a $224\times224$ RGB tensor with ImageNet channel normalization. Normalized box-center coordinates, widths, and heights from the YOLO-format proposal files were interpreted relative to the current source image. The same source dimensions were used to convert these coordinates into ROI crops. Each valid crop was resized to $80\times80$ pixels, normalized, and encoded separately.

Training transforms included horizontal reflection with corresponding updates to the horizontal box-center coordinates, color jitter for the global image, and color jitter for the lesion ROIs. The corresponding evaluation transforms omitted stochastic augmentation. Because preprocessing preceded ROI extraction, the archived implementation does not support the claim that local ROI pixels always came from unsegmented full-resolution photographs.

\subsection{Stage 1: Global Image Representation and Severity Prior}
\label{sec:global_prior}

The selected automatic and \texttt{NO\_CNN} oracle configurations used a global Vision Transformer stream from \texttt{timm}, initialized with \texttt{vit\_base\_patch16\_224.augreg\_in21k\_ft\_in1k}. The image features were projected into a 768-dimensional global representation and passed to a categorical head that produced four logits. Softmax converted these logits, $\mathbf{z}_g$, into the probability vector $\mathbf{p}_g$.

The fusion input also included four deterministic descriptors of $\mathbf{p}_g$: the largest class probability, the second-largest probability, their difference, and Shannon entropy, computed as $-\sum_c p_{g,c}\log(p_{g,c})$ with a numerical stabilizer. These descriptors summarize prediction concentration. They do not establish probability calibration or calibrated uncertainty under distribution shift.

The global ViT stream was distinct from the separately trained ConvNeXtV2-L comparator, which used an independent classifier checkpoint, preprocessing procedure, and prediction head. During training of the fused model, the global stream was frozen for the first three epochs and then allowed to receive gradient updates. The detector was trained separately. The term ``end-to-end'' therefore applies only to the differentiable components of the grading model; it does not imply joint optimization through YOLO proposal generation.

\subsection{Stage 2: Lesion Detection, ROI Embeddings, and Token Construction}
\label{sec:lesion_tokens}

The automatic pathway used a YOLO11s detector trained separately on ACNE04. It operated on a $960\times960$ detector canvas with an archived inference confidence threshold of 0.075 and a maximum of 16 selected boxes per image. Its weights were maintained separately from the grading network. The detector pipeline used an ACNE04-derived training split and its own validation procedure. Additional detector configurations and confidence thresholds were examined in retrospective comparison scripts. The available records do not establish that the final confidence threshold was selected independently of all test observations. Threshold-sweep results are therefore treated as exploratory rather than unbiased evidence for model selection.

For each image, $n$ was the number of parsed candidate boxes before top-$K$ truncation, and $K\leq16$ was the number of selected proposals. Boxes were ranked by confidence and retained up to the configured capacity. Each valid $80\times80$ ROI patch was processed by an ImageNet-pretrained ResNet-18, whose output was projected into a 192-dimensional appearance descriptor.

A six-component metadata vector contained the normalized box-center coordinates, width, height, detection confidence, and normalized count $\min(n,65)/65$. This vector was projected into 64 dimensions. The appearance and metadata descriptors were concatenated and mapped to a 256-dimensional lesion token. Proposal identity, confidence, and count were recorded separately from the reference clinical severity label.

In the annotation-assisted configuration, registered dermatologist boxes replaced detector proposals and received a confidence value of one for model input. The frozen external oracle protocol did not use annotated lesion-type identifiers as a separate categorical feature.

\subsection{Lesion Set Transformer and Count-Based Prior}
\label{sec:set_encoder}

The token encoder contained two self-attention blocks with four heads, feed-forward expansion, and padding masks. A learned aggregation token was prepended to the lesion tokens, and no lesion positional encodings were used. The final state of the aggregation token provided a 256-dimensional set summary.

For a fixed set of selected tokens, permuting the lesion tokens and their corresponding validity masks leaves the aggregation output unchanged:
\begin{equation}
    \mathbf{h}_L(P\mathbf{T},P\mathbf{m})
    =
    \mathbf{h}_L(\mathbf{T},\mathbf{m}),
    \label{eq:hl}
\end{equation}
where $P$ permutes only lesion-token positions. The archived numerical audit tested 100 such permutations, including empty and padded sets. This check concerns the order of the selected tokens. It does not establish invariance to upstream confidence ranking, top-$K$ selection, or detector nondeterminism. When no valid boxes were available, padded ROI tensors and the learned aggregation token allowed the encoder to produce a representation. An empty set did not automatically set the fusion gate to zero.

A ten-component proposal-statistics vector summarized count capacity, confidence moments, area moments, and coarse spatial occupancy. This vector was projected into 64 dimensions. Its final spatial component was implemented as the sum of the upper- and lower-half fractions. It was therefore constant for nonempty sets and did not provide an independent measure of vertical distribution.

The raw box count $n$ also defined a smoothed four-class prior using thresholds at 5, 20, and 50, with smoothing parameter $\alpha=0.08$:
\begin{equation}
    \boldsymbol{\pi}_n
    =
    (1-\alpha)\mathbf{e}_{q(n)}
    +
    \frac{\alpha}{4}\mathbf{1}_4,
    \qquad
    q(n)\in\{0,1,2,3\}.
    \label{eq:pi}
\end{equation}
Here, $\mathbf{e}_{q(n)}$ is the one-hot class vector obtained from the count thresholds. In automatic training, $n$ was the predicted proposal count; in oracle training, it was the annotation-box count. Neither count was verified to include only unilateral inflammatory lesions as required by Hayashi's diagnostic definition.

\subsection{Stage 3: Gated Residual Fusion and Conditional Identity Fallback}
\label{sec:grf}

The fusion input concatenated a 768-dimensional global embedding, a 256-dimensional lesion-set embedding, a 64-dimensional detection-summary embedding, four global logits, four global probabilities, four uncertainty descriptors, and a four-component count prior. The resulting 1,104-dimensional vector was mapped to a 768-dimensional hidden state through a layer-normalized projection. The residual head used widths of $768\rightarrow768\rightarrow4$, while the scalar gate head used widths of $768\rightarrow128\rightarrow1$ and a final sigmoid activation. Both heads included normalization, nonlinear activation, and dropout. These dimensions agree with the archived model definitions and the parameter shapes inspected in the selected checkpoints.

The final class logits and probabilities were computed as
\begin{equation}
    \mathbf{z}_f = \mathbf{z}_g + g\,\Delta\mathbf{z},
    \qquad
    g = \sigma\!\left(f_{\mathrm{gate}}(\mathbf{h}_F)\right),
    \qquad
    \mathbf{p}_f = \operatorname{softmax}(\mathbf{z}_f).
    \label{eq:zfinal}
\end{equation}
The identity $\mathbf{z}_f=\mathbf{z}_g$ holds when $g=0$ or $\Delta\mathbf{z}=0$. For finite inputs, the sigmoid ordinarily produces $0<g<1$, and the inspected automatic inference code did not force the gate to zero for empty proposal sets or atypical inputs. The formulation therefore provides a conditional mathematical identity. It does not establish detector-failure detection, reliability calibration, or clinical safety.

\begin{figure*}[t]
\centering
\includegraphics[width=\textwidth]{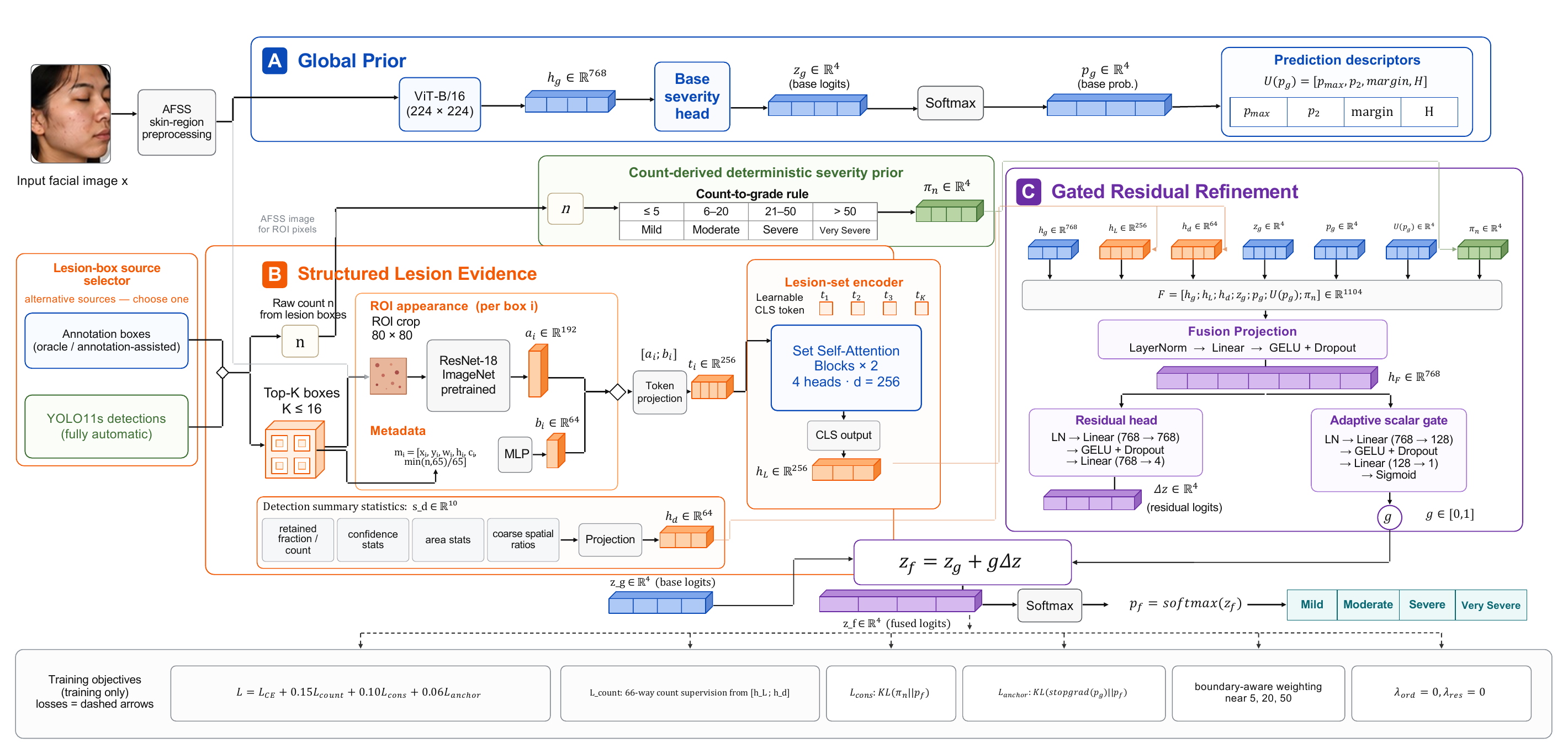}
\caption{LENS-GRF architecture, including the global ViT prior, alternative sources of lesion boxes, ROI and set encoding, count evidence, and gated residual refinement. The audited model uses a 1,104-dimensional fusion input and a 768-dimensional fusion hidden state. The annotation-box route is used only for evaluation as an alternative to automatic YOLO proposals. The fused logits equal the global logits when the gate or residual update is zero.}
\label{fig:framework}
\end{figure*}

\subsection{Training Objectives, Optimization, Model Selection, and Reproducibility}
\label{sec:training}

The executed automatic YOLO11s configuration and the selected ViT-only \texttt{NO\_CNN} oracle configuration used an objective combining categorical severity cross-entropy, count-bin classification, count-prior consistency, and divergence from the global prediction:
\begin{equation}
\begin{aligned}
    \mathcal{L}
    ={}&
    \mathcal{L}_{\mathrm{CE}}
    + 0.15\,\mathcal{L}_{\mathrm{count}}
    + 0.10\,D_{\mathrm{KL}}
    \left(\boldsymbol{\pi}_n\Vert\mathbf{p}_f\right) \\
    &+ 0.06\,D_{\mathrm{KL}}
    \left(\operatorname{stopgrad}(\mathbf{p}_g)\Vert\mathbf{p}_f\right).
\end{aligned}
\label{eq:ltotal}
\end{equation}

Severity cross-entropy used label smoothing of 0.03. The classification and count terms also used count-based sample weights near grade boundaries, defined as $1+0.35\exp[-d(n,\{5,20,50\})/2.5]$. The count head predicted 66 bins indexed from 0 to 65, with bin 65 used for overflow. Its target was the clipped input-box count: the detector-proposal count for the automatic model and the supplied annotation-box count for the oracle model. The automatic count target was therefore not an independent manual lesion tally.

The anchor term was the KL divergence from detached global probabilities, rather than an $L_2$ penalty on residual logits. The software included an ordinal auxiliary objective and a residual-norm penalty, but both received zero weights in the selected configurations. Other ablations used different combinations of active losses.

AdamW used learning rates of $8\times10^{-6}$ for the global stream, $2\times10^{-5}$ for ROI features, and $8\times10^{-5}$ for fusion and head parameters. Weight decay was 0.015. Training used four warm-up epochs, cosine learning-rate scheduling, and a maximum of 55 epochs. Mini-batches of four images with four-step gradient accumulation gave a nominal effective batch size of 16. Gradient norms were clipped at 1.0.

Training used a patience of 12 epochs, and the best checkpoint was selected using the validation score $0.40\,\mathrm{Acc}+0.25\,\mathrm{MacroF1}+0.35\,\mathrm{QWK}$. The automatic configuration used seed 42. Internal oracle sensitivity runs used seeds 42, 52, and 62. The detector was trained separately on the ACNE04-derived YOLO split using an SGD-based YOLO training schedule.

The external evaluation manifests record distinct fixed hashes for the independently trained global classifier, automatic fusion checkpoint, lesion detector, and selected oracle checkpoint. Replication requires these exact model assets together with the associated split, proposal, and registration files. The archived external run records Python~3.12.13, PyTorch~2.10.0+cu128, \texttt{timm}~1.0.26, OpenCV~4.13.0, and two NVIDIA Tesla T4 GPUs. These records describe the external inference environment; the corresponding training-environment versions were not independently confirmed.

\subsection{Comparators, Ablations, and Internal Evaluation Design}
\label{sec:comparators}

The study retained three distinct comparators: an independently trained ConvNeXtV2-L global classifier, the global ViT logits within LENS-GRF, and an XGBoost meta-fusion classifier supplied with ground-truth lesion counts. The count-assisted classifier is an oracle comparator and cannot be interpreted as an image-only clinical inference system. The principal automatic model used YOLO11s proposals, while the annotation-assisted model used ground-truth lesion boxes. These regimes differ in the information available to the model, even when they are evaluated against the same four image-level labels.

The experiment archive also included changes to gate use, global-branch composition, auxiliary objectives, and detector settings. Component-ablation scripts removed the sigmoid gate, replaced residual fusion with direct fusion, omitted uncertainty descriptors, compared CNN-only and ViT-only global branches, and changed loss weights. These runs were not all matched for backbone composition, annotation source, training budget, or seed.

Detector comparisons included YOLO12s, alternative confidence thresholds, and different maximum box counts. These retrospective experiments did not constitute a prespecified independent trial for threshold selection. Table~\ref{tab:methods_regimes} summarizes representative executed regimes and distinguishes automatic inference from annotation-assisted evaluation.

\begin{table*}[t]
\centering
\caption{Experimental comparison regimes and available lesion evidence. Sample counts and accuracy outcomes are omitted. The listed configurations do not imply equal information availability or training budgets.}
\label{tab:methods_regimes}
\small
\begin{tabularx}{\textwidth}{@{}p{2.3cm}p{2.05cm}p{2.2cm}p{2.35cm}X@{}}
\toprule
Configuration &
Global representation &
Lesion evidence &
Main comparison factor &
Evaluation unit and evidence scope \\
\midrule

ConvNeXtV2-L global &
Independent ConvNeXtV2-L &
None &
Global-image baseline &
ACNE04 image; PLSBRACNE01 subject \\

LENS internal prior &
ViT-B/16 within the fused model &
None used for base logits &
Global pathway within the fused model &
ACNE04 image; external subject \\

Automatic LENS-GRF &
ViT-B/16 &
YOLO11s, threshold 0.075, top 16 boxes &
Learned gate and residual update &
ACNE04 image, seed 42; frozen external subject \\

Automatic detector alternatives &
ViT-B/16 &
Alternative threshold, top-$K$, or YOLO12s &
Exploratory operating-point comparisons &
ACNE04 image; settings require identification of the individual run \\

GT oracle (ViT-only) &
ViT-B/16 &
Existing expert boxes &
Annotation-assisted set evidence &
ACNE04 image, three logged seeds; external matched subject across three raters, seed 42 \\

Oracle architecture/loss ablations &
Variant-specific &
Existing expert boxes &
Gate, branch, or loss removal &
ACNE04 image; primarily single-run comparisons \\

XGBoost count meta-fusion &
Separate global baseline &
GT lesion count &
Count-informed oracle comparator &
ACNE04 image; not autonomous inference \\

\bottomrule
\end{tabularx}
\end{table*}

\subsection{Frozen External Evaluation, Multi-View Inference, and Dermatologist-Annotation Registration}
\label{sec:external_protocol}

External evaluation used fixed ACNE04-trained global, detector, and fusion weights for all 200 PLSBRACNE01 participants. The recorded settings were a detector confidence threshold of 0.075, a $960\times960$ detector input, a maximum of $K_{\max}=16$ proposals, a $224\times224$ global input, and $80\times80$ ROI inputs. According to the frozen-run manifest, external images were not used for fine-tuning, threshold selection, checkpoint selection, or probability recalibration.

The frontal, left, and right views were processed independently. Their four-class probability vectors were averaged, and the class with the highest mean probability determined the participant's predicted grade. Reported probabilities for the internal ViT component came from the fusion model's global head, which was distinct from the separately trained ConvNeXtV2 classifier.

The expert lesion annotations were originally associated with facial crops stored under \texttt{image\_cut}. For annotation-assisted evaluation, these crops were registered to the full facial views using SIFT descriptors, two-nearest-neighbor matching, Lowe's ratio filter with a threshold of 0.75, and RANSAC homography estimation. Source and target working images were isotropically downscaled to a maximum dimension of 1,400 pixels. The estimated homography was converted to the original coordinate systems using the inverse target scaling matrix and the forward source scaling matrix.

A registration was accepted only if it had at least 12 matches, at least 10 geometric inliers, an inlier fraction of at least 0.45, and a median reprojection error of no more than three working-image pixels. The estimated scale also had to lie between 0.5 and 2.0, with an absolute rotation of no more than $15^\circ$. Valid transformed boxes were normalized relative to the dimensions of the corresponding full facial view.

Registration eligibility was determined before annotation sources were compared. The archived manifest identified 450 eligible crop views with unambiguous associations, representing 150 participants. Complete registration across all three views was available for 149 participants. Requiring usable, matched annotations from Doctors~A, B, and C reduced the cohort to 148 participants.

Oracle-versus-automatic comparisons therefore used the same 148 participants and 444 views, while the full-cohort automatic evaluation retained all 200 participants. The matched cohort was selected through registration and annotation requirements and was not a random sample of the full external population. The three dermatologists supplied lesion bounding boxes. The analysis measured agreement among model predictions generated from their respective annotations, not direct agreement among independently assigned Pillsbury severity grades.

\subsection{Evaluation Metrics, Statistical Tests, and Evidence Boundaries}
\label{sec:statistics}

ACNE04 image-level predictions and PLSBRACNE01 subject-level predictions were evaluated as four ordered categories. Accuracy was the proportion of predictions matching the reference grade. Per-class precision was calculated as $TP/(TP+FP)$ and sensitivity, or recall, as $TP/(TP+FN)$. Specificity was calculated for each grade against all remaining grades as $TN/(TN+FP)$. Macro precision, macro sensitivity, and macro specificity were the arithmetic means of the corresponding four class-level measures.

Macro-F1 was the mean of the class-level F1-scores, each calculated as the harmonic mean of precision and recall. Balanced accuracy averaged class-level recall and therefore equaled macro sensitivity under the same class definitions. The archived scoring code assigned zero to undefined class-level precision. The treatment of other undefined denominators was documented during supplementary recomputation.

Ordinal mean absolute error measured the integer distance between predicted and reference grades. Quadratic weighted kappa assigned larger penalties to errors spanning more grades, while the multiclass Brier score used the full predicted probability vector. Where reported, negative log-likelihood and calibration error were secondary descriptive measures and did not replace external calibration testing.

Paired subject-level comparisons used exact two-sided McNemar tests based on whether each participant was classified correctly by each method. Holm's step-down adjustment was applied to the nine comparisons in the matched three-oracle analysis. The separate full-cohort automatic-versus-global analysis used its recorded two-comparison family. Unadjusted results were distinguished from family-adjusted inference, and post-hoc operating-point comparisons were not treated as prespecified confirmatory tests.

Nonparametric percentile 95\% confidence intervals were calculated from 5,000 bootstrap samples drawn at the subject level with replacement. The three correlated views from each participant were not treated as independent observations. The frozen full-cohort and matched-oracle manifests recorded bootstrap seeds 20260901 and 20260902, respectively.

Where necessary, metrics were recomputed from saved subject-level reference grades, averaged class probabilities, and paired predictions. Reported differences apply to the specified cohorts and the correspondence used between grading indices. They do not isolate the effects of image-acquisition shift from those of grading-criterion shift or establish prospective clinical utility.